%% file: sample-sigconf.tex
\documentclass[sigconf]{acmart}

\usepackage{booktabs} 
\usepackage{amsmath}
\usepackage{amsfonts}
\usepackage{multirow}
\usepackage{graphicx}
\usepackage{url}

\setcopyright{rightsretained}

\setcopyright{acmlicensed}
\acmConference[DAPA '19]{DAPA 2019 WSDM Workshop on Deep matching in Practical Applications}{February 15th, 2019}{Melbourne, Australia} 
\acmYear{2019}
\copyrightyear{2019}
\acmDOI{}
\acmISBN{}
\acmPrice{}

\begin{document}
\title{Dynamic Feature Generation Network for Answer Selection}

\begin{abstract}
Extracting appropriate features to represent a corpus is an important task for textual mining. Previous attention based work usually enhance feature at the lexical level, which lacks the exploration of feature augmentation at the sentence level. In this paper, we exploit a Dynamic Feature Generation Network (DFGN) to solve this problem. Specifically, DFGN generates features based on a variety of attention mechanisms and attaches features to sentence representation. Then a thresholder is designed to filter the mined features automatically. DFGN extracts the most significant characteristics from datasets to keep its practicability and robustness. Experimental results on multiple well-known answer selection datasets show that our proposed approach significantly outperforms state-of-the-art baselines. We give a detailed analysis of the experiments to illustrate why DFGN provides excellent retrieval and interpretative ability.
\end{abstract}

%
%



\maketitle

\input{samplebody-conf}

\bibliographystyle{ACM-Reference-Format}
\bibliography{bibliography}

\end{document}

%% file: samplebody-conf.tex
\section{Introduction}
Answer sentence selection is a sub-task of question answering (QA) and a popular topic of information retrieval (IR) in the past years. Answer selection includes factoid QA seeking out facts and non-factoid QA choosing complicate answer texts. One public factoid dataset in answer selection is WikiQA\footnote{http://aka.ms/WikiQA/}, we give an example of a question with a positive answer and four negative answers extracted from WikiQA dataset in Table.\ref{QAexample}. The goal of answer selection is to select correct answers from five answer candidates.

\begin{table}[htbp] 
\caption{Example of answer selection in WikiQA dataset.}
\label{QAexample}
\begin{center}
\begin{tabular}[b]{|l|}
\hline
\textbf{Question:} how are glacier caves formed ?   \\ \hline
\textbf{Positive answer:} \\
A glacier cave is a cave formed within the ice of a glacier.  \\ \hline
\textbf{Negative answers:} \\
\textbf{1.} A partly submerged glacier cave on Perito Moreno Glacier. \\
\textbf{2.} The ice facade is approximately 60 m high. \\
\textbf{3.} Ice formations in the Titlis glacier cave. \\
\textbf{4.} Glacier caves are often called ice caves , but this term \\
is properly used to describe bedrock caves that contain \\
year-round ice.\\
\hline
\end{tabular}
\end{center}
\end{table}

Comprehending logical and semantic relationship between two sentences and acquiring the ability to rank is essential in answer selection task. In recent years, the attention mechanism, allocating different weight to words or sub-phrases in sentences~\cite{bahdanau2015neural,rocktaschel2016reasoning}, is widely employed in natural language processing. Searching appropriate semantic information for ranking answers naturally conforms to the characteristics of attention mechanism. By computing word level similarity matrix to extract matching patterns from different textual granularity that is useful for prediction, attention significantly improve the performance of convolutional~\cite{yin2016abcnn:,he2016pairwise} or recurrent~\cite{wang2016learning,wang2017bilateral} networks. Compare-aggregate models with soft-attention~\cite{parikh2016a,wang2017a} provide a new way of comparing attention results and obtain the impressive achievement. Soft-attention considers all the words in a sentence when assigning weights, also known as word-by-word attention. Intra-attention based models~\cite{liu2016learning,lin2017a} extract relationship between words in a sentence. Co-attention based models~\cite{gong2018natural,kim2018semantic,tran2018multihop} learn joint information with respect to two sentences then distribute weight to both sides. Attention mechanism comes in different forms and in company with extractive max and mean-pooling~\cite{santos2016attentive,zhang2017attentive} or alignment pooling~\cite{shen2017inter-weighted,chen2017enhanced}. Extractive max-pooling selects each word based on its maximum importance of all words in the other text. Extractive mean-pooling is a more wholesome comparison, paying attention to a word based on its overall influence on the other text. Alignment-pooling aligns semantically similar sub-phrases together, extracting only the most relevant information. Although the attention mechanism is typically applied as a weight allocation strategy, recent approaches~\cite{tay2018a,tay2018multi-cast} utilize attention as feature augmentation tool. They utilize co-attention and intra-attention in conjunction with extractive max, mean, alignment pooling to compute weighted representations, then compress each representation into a scalar, attaching these scalars as extra features to original embedding. Their methods are proven to be efficient in retrieval-based question answering tasks. 

However, previous approaches have two disadvantages. The first is that they only focus on the augmentation of lexical level~\cite{shen2017inter-weighted,chen2017enhanced,tay2018a,tay2018multi-cast}, while research on feature enhancement at sentence level is not enough. We believe that strengthening overall characteristics is more consistent with the habits of human cognition. Therefore, in this paper, we first extract and enhance features at the sentence level and then perform information retrieval process. The second deficiency is that previous studies lack dynamic threshold mechanism when filtering features. Alignment results are usually directly employed~\cite{shen2017inter-weighted} or compressed into scalars~\cite{tay2018multi-cast}. DCA~\cite{bian2017a} use dynamic clipping methods, but their thresholds are still empirical values. In this work, we apply parametric co-attention to perform thresholds in filtering operation. Different kinds of characteristics are automatically distilled according to the specific corpus.

The main contributions of our work are as follows:
\begin{itemize}
\item We propose a sentence level feature enhancement method, extend feature augmentation method that used to focus on the word level. In addition to the traditional attention algorithms, we propose a new abstract feature extraction algorithm and a new dynamic threshold algorithm for feature selection. Our method requires no additional information and relies solely on the original text.
\item We design Dynamic Feature Generation Network for answer selection task, DFGN acquires the ability to automatically abandon useless and inefficient data, reducing the cost of adjusting parameters in the neural network. 
\item  Experimental results show that DFGN outperforms current work on WikiQA, TREC-QA and InsuranceQA datasets. We give an in-depth analysis to illustrate why DFGN owns the excellent retrieval and interpretative abilities. Our code is publicly available\footnote{https://github.com/malongxuan/QAselection}.
\end{itemize}

\section{Related Work}
Learning to rank candidate answers is a long-standing problem in NLP and IR. The dominant state-of-the-art models today are mostly neural attention based approaches. These models focus on different aspects. MPCNN~\cite{rao2016noise-contrastive} proposes a new ranking method in answer selection. Pairwise PWIM~\cite{he2016pairwise} combining LSTMs and deep CNN to investigate the similarity matrix. Gated attention in RNN such as~\cite{rocktaschel2016reasoning} and~\cite{wang2016learning} explore the internal semantic relations of sentences, obtain remarkable improvement in natural language inference. IABRNN~\cite{wang2016inner} further develop gated attention and achieve success in answer selection. Inner-attention~\cite{liu2016learning} and self-attention~\cite{lin2017a} are introduced to LSTM, extracting an interpretable sentence embedding. 

Recent advances in neural matching models go beyond independent expression learning. Major architectural paradigms that invoke interaction between document pairs directly improve performance because matching has deeper and finer granularity. Multi-Perspective CNN~\cite{he2015multi-perspective} and BiMPM~\cite{wang2017bilateral} match sentences with multiple views and perspectives. Recurrent neural network model such as AP-BiLSTM~\cite{santos2016attentive}  utilize extractive max-pooling to learn the relative importance of a word based on its maximum importance to all words in the other document. IWAN~\cite{shen2017inter-weighted} extract features from a constructed word-by-word alignment matrix with self-attention. We use all previous extraction strategies and design a novel fixed attention features with parameterized self-attention.

External resources are also used for sentence matching. Syntactic parse trees or WordNet ~\cite{chen2017enhanced,chen2018natural} are employed to measure the semantic relationship among the words. Unlike these, we do not use any external resources. Meanwhile, Compare-aggregate framework~\cite{parikh2016a,wang2017a} are proven to be effective in answer selection tasks. DCA~\cite{bian2017a} further improves former work by dynamic clipping useless information, which inspires our feature filtering method. 

However, the approaches using deeper layers lead to more progress in performance. MAN~\cite{tran2018multihop} applies multihop-sequential-LSTM to achieve step by step learning. Most recently, Compare-propagate model CAFE~\cite{tay2018a} and multi-cast approach MCAN-FM~\cite{tay2018multi-cast} adopt efficient compression of attention vectors into scalar valued features, which are used to augment the base word representations, develop another way of employing attention, inspired by them, we excavate the joint approach which uses multiple attentions and extractive pooling strategies to enhance representations power for QA. But unlike all previous work, we design dynamic feature generation and selection methods at sentence levels.

\section{Our Proposed Model}
\begin{figure}[!thb]
  \centering
 \includegraphics{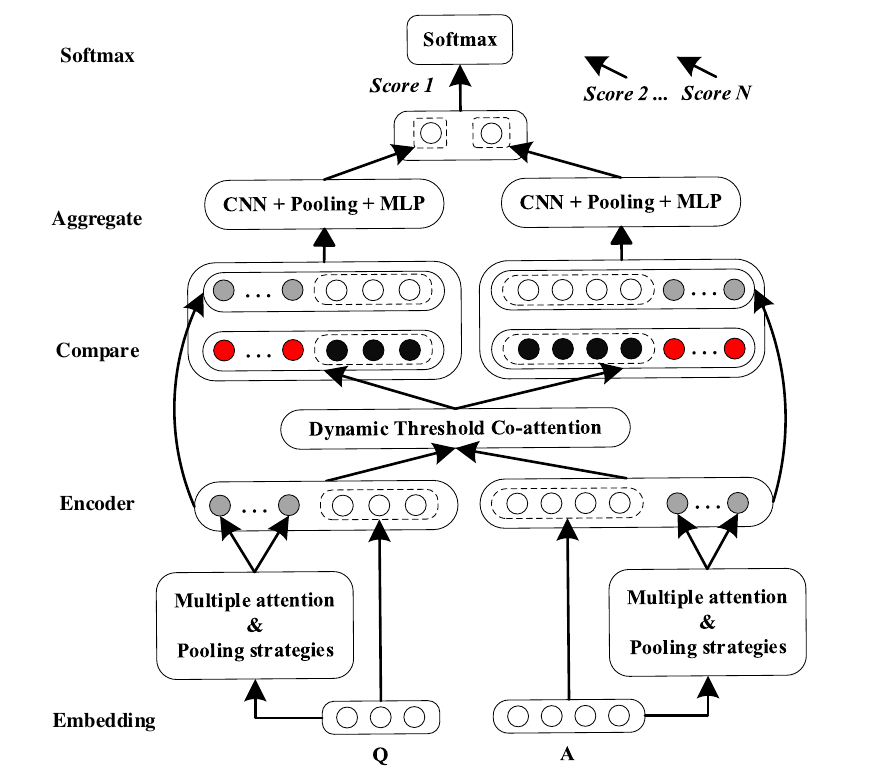}
\caption{The over structure of DFGN.}
\label{full_model}
\end{figure}

The overall structure of DFGN is shown in Figure \ref{full_model}. The inputs of our model are query and answer sentences. We apply pre-trained $300$ dimensional Glove~\cite{pennington2014glove:} as word embedding. $Q\in \mathbb{R}^{q \times d}$ and $A\in \mathbb{R}^{a \times d}$ are initial inputs, $d$ represents the $300$ embedding size, $q$ and $a$ represent the length of $Q$ and $A$ respectively.  Next we will portray this model in detail.

\begin{table*}[htbp]
\centering
  \caption{Statistics of WikiQA, TREC-QA and InsuranceQA datasets.}
  \label{table_data}
  \begin{tabular}{l|c|c|c}
    \hline
     Dataset(train/test/dev) &WikiQA &TrecQA(clean) &InsuranceQA V1 \\
    \hline
Questions&873/ 243/126 & 1162/68/65 &12887/1800$\ast$2/1000 \\
Sentences& 20360/6165/2733      &5919/1442/1117 &24981(ALL)   \\
Average length of questions&7.16/7.26/7.23  &11.39/8.63/8.00  &7.16        \\
Average length of sentences&25.29/24.59/24.59 &30.39/25.61/24.9 &49.5      \\
Question answer pairs         &5.9k/1.4k/1.1k      &  53.4k/1.4k/1.1k   &1.29m/1.8m/1m   \\
Average candidate answers  &9              &38         &100/500/500          \\
    \hline
  \end{tabular}
\end{table*}

\subsection{Multiple Attention and Pooling Strategies}
Co-attention and intra-attention are the most commonly used attention mechanisms. The traditional co-attention usually compute the affinity matrix between $Q$ and $A$ using $Q \otimes A^T$, $Q \otimes W \otimes A^T$ and $W \otimes [Q;A]^T$, etc. with activation function such as $tanh$, $sigmoid$ and $relu$, etc. $Q$ and $A$ are sentences representations, $W$ is parameter matrix, $\otimes$ means matmul product, $[;]$ is concatenation. The intra-attention matrices are usually calculated as $Q \otimes Q^T$, $Q \otimes W \otimes Q^T$ and $W \otimes Q^T$. In our model, we use both parametric and non-parametric computations to extract intrinsic features while learning extended features with parameters, we define self-attention and intra-attention respectively for the sake of distinction. Self-attention matrices $S^Q\in \mathbb{R}^{q \times d}$ and $S^A\in \mathbb{R}^{a \times d}$, intra-attention affinity matrices $I^Q\in \mathbb{R}^{q \times q}$ and $I^A\in \mathbb{R}^{a \times a}$, co-attention affinity matrices $C^Q\in \mathbb{R}^{a \times q}$ and $C^A\in \mathbb{R}^{q \times a}$ are calculated as follows, where $\delta$ means $Tanh$, $W^{S^Q}$ and $W^{S^A}\in \mathbb{R}^{d \times d}$ are parameters. %

\begin{align}
S^Q = \delta(Q \otimes W^{S^Q})\ &\&\ S^A = \delta(A \otimes W^{S^A})  \\
I^Q = Q \otimes Q^T\ &\&\ I^A = A \otimes A^T  \\
C^Q = A \otimes Q^T\ &\&\ C^A = Q \otimes A^T
\end{align}

We extract feature vectors in a variety of ways then concatenate them to sentence representations. The features are extracted from three levels: word interaction between sentences, word interaction within sentences, and the hidden layer interaction of sentence. They are corresponding to co-attention, intra-attention, self-attention respectively. We apply extractive max pooling to $S^Q$ as formula $(4)$ to obtain $M^{S^Q}$. 

\begin{align}
M^{S^Q} = \sum^{q}_{i=1}{\frac{ exp(\max \limits^{d}_{j=1}S^Q_{i,j}) }{\sum^{q}_{t=1}exp(\max \limits^{d}_{j=1}S^Q_{t,j}) }}\ast{Q_i} 
\end{align}

Where $M^{S^Q}\in \mathbb{R}^{d \times 1}$, $\ast$ is element-wise product, and analogously for $M^{S^A}, M^{I^Q}, M^{I^A}, M^{C^Q}$ and $M^{C^A}$. It is worth noting that $M^I$ and $M^C$ extract the sentence level dimensional features, while $M^S$ extracts the hidden dimensional features. Employing extractive pooling to hidden dimension in answer selection is the first time to our knowledge. Extractive mean-pooling results, denoted by $N$, are calculated similarity as $M$ with $max$ replaced by $mean$, we give the formula of $N^{S^Q}\in \mathbb{R}^{d \times 1}$ as follows. 

\begin{align}
N^{S^Q} = \sum^{q}_{i=1}{\frac{ exp(\frac{1}{d}\sum \limits^{d}_{j=1}S^Q_{i,j}) }{\sum^{q}_{t=1}exp(\frac{1}{d}\sum \limits^{d}_{j=1}S^Q_{t,j}) }}\ast{Q_i} 
\end{align}

It is similar for $N^{S^A}, N^{I^Q}, N^{I^A}, N^{C^Q}$ and $N^{C^A}$. Now we get six feature vectors for $Q$ and $A$ respectively. Then we employ alignment pooling to $I$ and $C$, the weighted representation $R^{I^Q}$ and $R^{C^Q}$ are calculated as formulas $(6)$ and $(7)$, and analogously for $R^{I^A}$ and $R^{C^A}$. Where $R^{I^Q}$ and $R^{C^Q}\in \mathbb{R}^{q \times d}$, and analogously for $R^{I^A}$ and $R^{C^A}$. Then we apply ordinary max-pooling and mean-pooling to $R^I$ and $R^C$. The max-pooling results denoted by $K^{I^Q}, K^{I^A}, K^{C^Q}, K^{C^A}\in \mathbb{R}^{d \times 1}$. The mean-pooling results denoted by $L$. 

\begin{align}
R^{I^Q}_j &= \sum^{q}_{i=1}{{\frac{ exp(I^{Q}_{i,j}) }{  \sum^{q}_{t=1}exp(I^{Q}_{t,j}) }}\ast {Q_i}}\\
R^{C^Q}_j &= \sum^{a}_{i=1}{{\frac{ exp(C^{Q}_{i,j}) }{  \sum^{a}_{t=1}exp(C^{Q}_{t,j}) }}\ast {A_i}}
\end{align}

So far, we have gained $10$ characteristics for $Q$ and $A$ respectively. By attaching feature vectors to $Q$, we obtain final sentence representation $X^Q = [Q$; $M^{S^Q}$; $M^{I^Q}$; $M^{C^Q}$; $N^{S^Q}$; $N^{I^Q}$; $N^{C^Q}$; $K^{I^Q}$; $K^{C^Q}$; $L^{I^Q}$; $L^{C^Q}]$$\in \mathbb{R}^{d \times (q+10)}$, and analogously for $X^A\in \mathbb{R}^{d \times (a+10)}$. 

\subsection{Encoder Layers}

\begin{table*}[htb]
\centering
\caption{Performance for answer sentence selection on WikiQA, TREC-QA and InsuranceQA datasets.}
\label{table_all}
\begin{tabular}{l|c|c|c|c|c}
\hline
\multirow{2}{*}{Models} & \multicolumn{2}{c|}{WikiQA} & \multicolumn{2}{c|}{TrecQA(clean)} & \multicolumn{1}{c}{InsuranceQA V1} \\
\cline{2-6}
& MAP & MRR & MAP & MRR &Top1(Test1/Test2)\\
\hline
AP-BiLSTM (~\cite{santos2016attentive}) & 0.671 & 0.684  & 0.713& 0.803  & 0.717/0.664 \\
MP-CNN (~\cite{he2015multi-perspective}) & 0.693 & 0.709 & 0.777 & 0.836  & - / - \\
MPCNN+NCE (~\cite{rao2016noise-contrastive}) & 0.701 & 0.718 & 0.801 & 0.877 & - / - \\
PWIM(~\cite{he2016pairwise})          &0.709&0.723    &-&-     &- / -\\
BiMPM (~\cite{wang2017bilateral}) & 0.718 & 0.731    & 0.802 & 0.875    & - / -\\
MS-LSTM(~\cite{tran2018multihop}) & 0.722 & 0.738    & 0.813 & 0.893    & 0.705/0.669 \\
IWAN (~\cite{shen2017inter-weighted}) & 0.733 & 0.750 & 0.822 & 0.889   & - / -\\
IABRNN (~\cite{wang2016inner}) & 0.734 & 0.742 & - & -      & 0.701/0.651 \\
MULT (~\cite{wang2017a}) & 0.743 & 0.754       & - & -     & \textbf{0.752}/\textbf{0.734}\\
DCA (~\cite{bian2017a}) & \textbf{0.756} & \textbf{0.764} &0.821 & 0.899    & - / - \\
MCAN-FM(~\cite{tay2018multi-cast}) & - & - &\textbf{0.838} & \textbf{0.904}  & - / -\\
\hline
DFGN(reduce)     & 0.745 &0.753   & 0.828&0.905     & 0.762/0.744\\
DFGN(full)      &\textbf{0.766}&\textbf{0.780}&\textbf {0.848}&\textbf{0.928}    & \textbf{0.775}/\textbf{0.757}\\
\hline
\end{tabular}
\end{table*}

After multiple attention, we use nonlinear functions as encoder to compute the input of the second co-attention layer separately, denoted by $E^Q\in \mathbb{R}^{d\times (q+10)}$, $E^A\in \mathbb{R}^{d\times (a+10)}$. Where $\sigma$ represent $Sigmoid$ and $\delta$ means $Tanh$. Experiments show that this encoding method not only achieves equal level accuracy as other complex structures like CNN and LSTM, but also has fewer parameters and saves training time. We present the formula of computing $E^Q$ as equation $(8)$, and analogously for $E^A$. $W^{X^Q}_{en1}$ and $W^{X^Q}_{en2}$ are parameters to be learned.

\begin{align}
E^Q &= \sigma(X^Q\otimes W^{X^Q}_{en1})\ast \delta(X^Q\otimes W^{X^Q}_{en2})
\end{align}

\subsection{Second Attention Layer}
The second co-attention layer learns joint information between features to perform interactive confirmation. We use the same formula as $(3)$ to calculate the affinity matrix $G^{E^Q}$. Then we apply parametric computation of co-attention as equation $(10)$ to get $\widetilde{G}^{E^Q}$ which perform thresholds, $\widetilde{W}^Q$ is a parameter matrix to be learned. We employ equation $(11)$ to get weighted representation $H^{E^Q}$, $\phi(x,y)$ is an operation that if $x<y$ then $x$ set to $0$, otherwise $x$ keeps original values. The advantage of this method is that it can remove redundant information dynamically instead of using empirical values. For different corpora, the number of retained features are different. The calculations are analogously for $G^{E^A}$, $\widetilde{G}^{E^A}$  and $H^{E^A}$. 

\begin{align}
G^{E^Q} &= E^A \otimes (E^Q)^T\\
\widetilde{G}^{E^Q} &= E^A \otimes \widetilde{W}^Q \otimes (E^Q)^T\\
H^{E^Q}_j &=\sum^{a}_{i=1} \phi(  \frac{ exp(G^{E^Q}_{i,j}) }{ \sum^{a}_{t=1}exp(G^{E^Q}_{t,j}) } ,\nonumber\\&\qquad\qquad {\frac{ exp(\widetilde{G}^{E^Q}_{i,j}) }{  \sum^{a}_{t=1}exp(\widetilde{G}^{E^Q}_{t,j}) }})\ast {E^A_i} 
\end{align}


\subsection{Compare, Aggregate, Softmax Layers}

There are several different forms of compare function~\cite{wang2017a}. We choose element-wise multiplication. The goal of the comparison layer is to match each feature with its weighted version. The comparison result, $Y^E = H^E\ast E$, is fed to CNN with both max-pooling and mean-pooling to get aggregate features, then we employ multilayer perceptron to get the final scores of $Q$ and $A$, then two branches are concatenated and compressed to a scalar for the final softmax layers, as shown in Figure \ref{full_model}. We feed related answer set $A$\{$A_1, A_2,\ldots, A_N$\}, target label set $Y$\{$y_1,  y_2, \ldots, y_N$\} along with $Q$ into the model. We select all positive answers to this question, denoted by $p$, then randomly select $N-p$ negative answers from the answer pool. We train our listwise model with KL-divergence loss to optimize the ranking results. Please consult our code to see the implementation.

\section{Experiments}

\subsection{Datasets and Experimental Protocol}

Statistical information of experimental datasets is shown in Table \ref{table_data}. We resort to Mean Average Precision (MAP), Mean Reciprocal Rank (MRR) and accuracy (Precision@1) to measure the experimental results. We use the pre-trained $300$ dimensional Glove vectors\footnote{https://nlp.stanford.edu/projects/glove/} to initialize our word embedding. We fix the word representations during training for a fair comparison. Due to self-learning ability, DFGN requires very few protocol changes between different datasets. 

WikiQA~\cite{yang2015wikiqa:} is a popular benchmark dataset for open-domain, factoid question answering. It is constructed by crowd-sourcing through sentences extraction from Wikipedia and Bing search logs. To train our model in mini-batch, we truncate the question to $12$ words, the answer to $50$ words, candidate answers to $15$ and batch size to $11$. We add $0$ at the end of the sentence if it is shorter than the specified length. We resort to Adam algorithm as the optimization method and update the parameters with the learning rate as $0.001$, $beta_1$ as $0.9$, $beta_2$ as $0.999$. The CNN windows are $[1,2,3,4,5]$. We set a dropout rate as $0.1$ at encoder layer. We add $L2$ penalty with the coefficient parameter $\lambda$ as $10^{-5}$. What’s more, in order to avoid the gradient exploding problem, we resort to the gradient global norm clipping method, setting the clip norm as $5$.

TREC-QA~\cite{wang2007what} is a well-known benchmark dataset collected from the TREC Question Answering tracks. Previous work~\cite{he2016pairwise,wang2016inner} used the raw version that has $82$ questions in the development set and $100$ questions in the test set. More recent work~\cite{shen2017inter-weighted,bian2017a,tay2018multi-cast} used clean version by removing questions that have only positive/negative answers or no answers, resulting in only $65$ questions in the validation set and $68$ questions in the test set. We use the clean version. The experiment settings are the same as WikiQA.

InsuranceQA is an exclusive domain, non-factoid answering dataset proposed by~\cite{feng2015applying}, collected from a community question answering website which contains two versions ($V1$ and $V2$). The average length of sentences in this corpus is $49.5$. For each question in train set, there are $100$ randomly selected candidate answers, for each question in the validation set and the test set, there are $500$ candidate answers. The longer average sentence length and more candidate answers make InsuranceQA more difficult. In this work, we use the $V1$ version which has two test sets, denoted by Test1 and Test2. 

We implement our model with Keras\footnote{https://github.com/keras-team/keras} based on Theano library\footnote{http://www.deeplearning.net/software/theano/}. Moreover, all experiments are carried out on Ubuntu 16.04, a single GPU (GeForce GTX 1080).

\subsection{Experimental Results}

Table \ref{table_all} reports our experimental results on all datasets. We selected $11$ models for comparison, and we re-implemented several of these models. However, it may be due to the problem of adjusting parameters, the accuracy of our redesigned model is generally lower than that in original papers proposed the corresponding model. Therefore, we mainly use the original papers data to compare . DFGN(reduce) represents a model without enhanced sentence features, and DFGN(full) means full model. On WikiQA, we observe that DFGN(full) outperforms a myriad of complex neural architectures. Notably, we obtain a clear performance gain of $1.6\%$ in terms of MRR against strong models such as MULT and DCA. Table \ref{table_all} also reports our results on the clean version of TrecQA. On the MRR index, DFGN(full) model outperforms latest MCAN by $2.4\%$. On InsuranceQA V1 dataset, our models outperform even the strongest model SUBMULT+NN~\cite{wang2017a}. DFGN(full) gains accuracy promotion of $2.3\%$ both in test1 and test2. In all experiments, DFGN(full) model achieve better retrieval performance than DFGN(reduce) model.

\section{DISCUSSION AND ANALYSIS}

\begin{figure}[!thb]
\centering
 \includegraphics[width=3in]{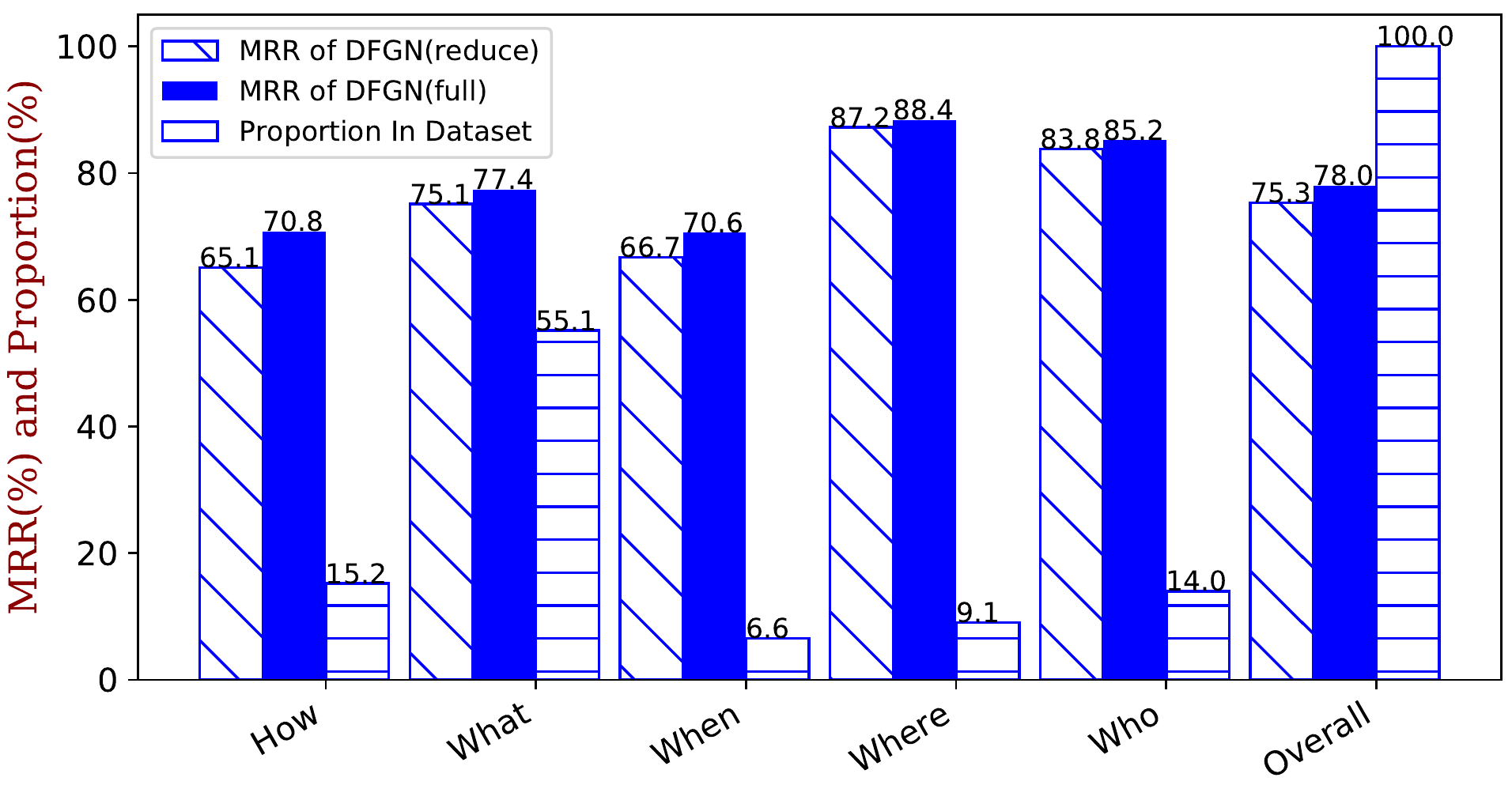}%
\caption{Comparison between fixed and dynamic features on type of questions.}
\label{ques_type}
\end{figure}

\begin{figure*}[htb]
  \centering
 \includegraphics[width=5.2in]{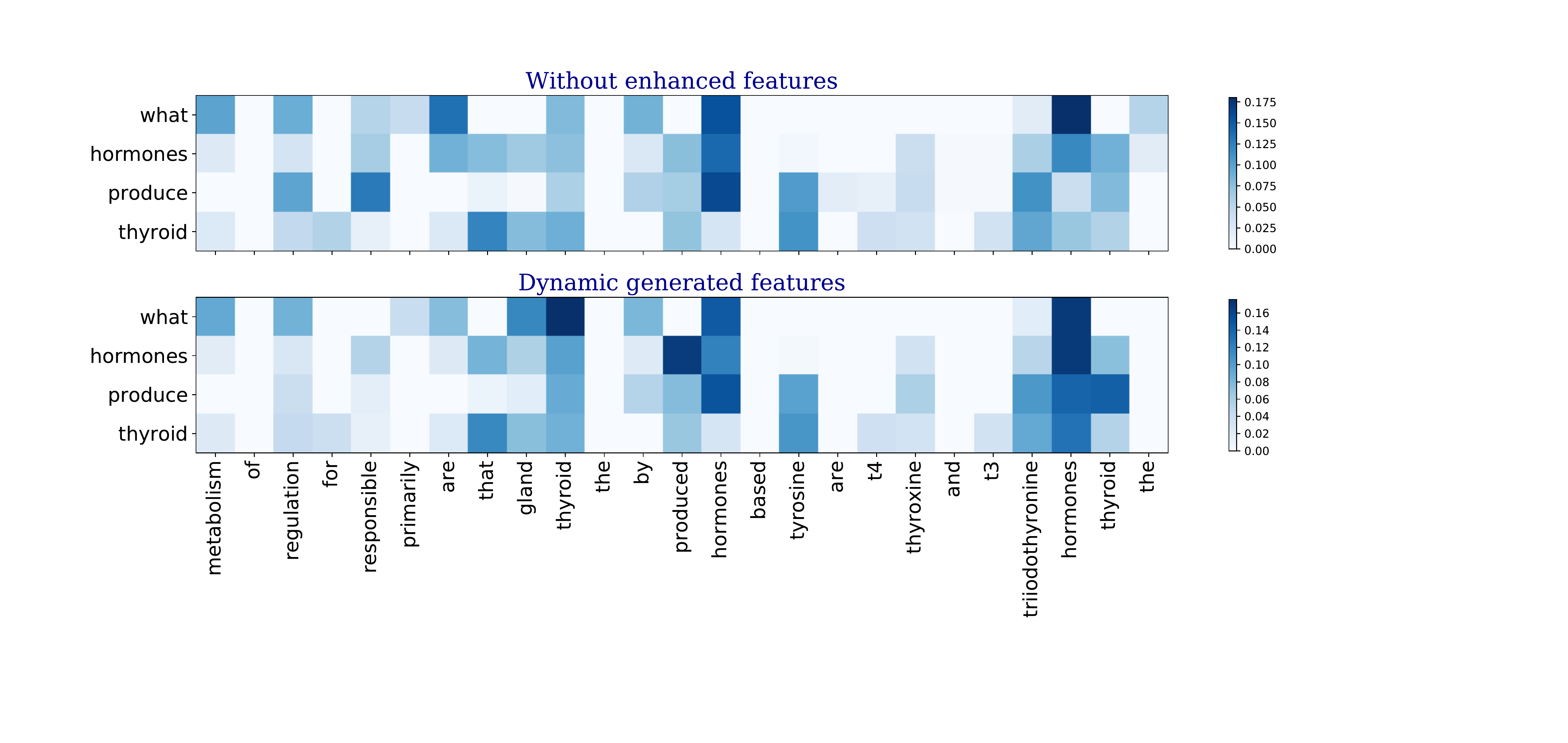}%
\caption{Comparison between No enhanced features and Dynamic attention features in second co-attention.}
\label{b_a}
\end{figure*}

\begin{figure}[htb]
  \centering
 \includegraphics[width=3in]{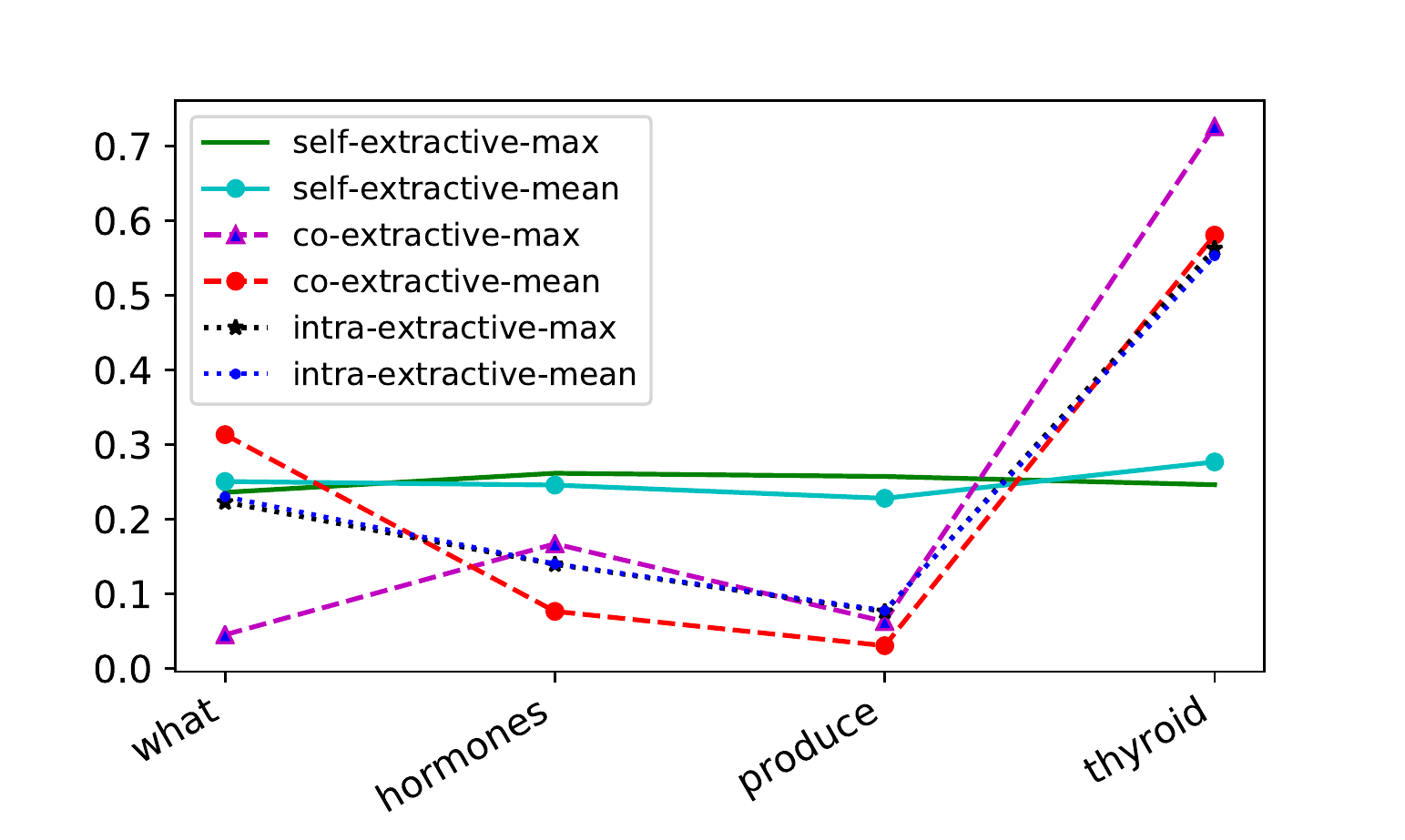}
\caption{Comparison between extractive pooling with co/intra/self attention.}
\label{pooling}
\end{figure}

\subsection{Question Type Analysis}

In this paragraph, we analyze two problems, first is the efficiency of the model itself for different types of problems, second is the impact of dynamic features on different problems. Figure \ref{ques_type} demonstrates all five types of questions in WikiQA test set. The histogram represents the MRR metric and the proportion of each type of questions in dataset respectively. We use DFGN(reduce) and DFGN(full) models to compare the difference. 

We can observe that both models own better results for 'Where' and 'Who' questions because locations and characters are easier to retrieve. In DFGN(reduce) model, the MRR value of $87.2\%$ and $83.8\%$ are respectively achieved for 'Where' and 'Who' questions. While due to the proportion of $55.1\%$ and $15.2\%$, the effect on 'What' and 'How' questions decide the overall performance of the model. In DFGN(reduce) model, the MRR value $75.1\%$ and $65.1\%$ are respectively achieved for 'What' and 'How' questions. After adding dynamic sentence features, we find that DFGN(full) model improves the MRR value by $2.3\%$ on 'What' questions, polish up the MRR value of 'How' question by $5.7\%$. 'What' and 'How' problems have increased more than the 'Where' and 'Who' problems, which shows that the dynamic feature generation successfully improves the comprehension ability of complex semantic relations.

\subsection{In-depth Analysis}

We resort to a question-answer pair of WikiQA dataset to perform in-depth analysis. Figure \ref{b_a} illustrates how sentence features affect the weight allocation of the second co-attention matrix. Since the second co-attention matrix in DFGN(full) owns $10$ additional features both in rows and columns, we only compared the common parts. The left heat map is from DFGN(reduce) model, and the right heat map is from DFGN(full) model. The question is "what hormones produce thyroid" and the answer is "the thyroid hormones triiodothyronine t3 and thyroxine t4 are tyrosine-based hormones produced by the thyroid gland that are primarily responsible for regulation of metabolism". It should be noted that this example was deliberately chosen by us. Because WikiQA dataset is collected from Wikipedia and Bing search logs, this ambiguous question is likely to arise when natural language questions are asked. What the questioner really wants to ask is "Which gland produces thyroid hormones". We use this example to show how ambiguous natural language questions are processed within our model.

In Figure \ref{b_a}, the left attention matrix put too much weight on irrelevant words or sub-phrase such as "responsible". In contrast, the right matrix with dynamic generated features better understand the meaning of the question and focus on a more appropriate word or sub-phrase such as "hormones" and "thyroid gland", meanwhile, it reduces the weight distribution of unimportant words such as "responsible" and "regulation". This is because after adding additional features, the weight distribution range changes from the original sentence length to the sentence length plus $10$. Since the relevant weights are improved, there is less noise in the attention matrix. Thus the retrieval result is more precise. Figure \ref{b_a} proves that by adding more features, it corrections the deficiencies of the original attention mechanism. 

We also present the extractive pooling attention weight in Figure \ref{pooling}, including co-attention, self-attention, and intra-attention. Co-attentions pay more attention to the core words, such as "thyroid". Self-attention assigns weights more evenly. Intra-attention pays more attention to interrogative word such as "what". Even in a short sentence, different attention mechanisms focus on different positions. When they act synthetically, the weight of ambiguous words in natural language questions is reduced, such as 'hormones' in this case. In practice, the longer the sentence is, the more words are related to the overall semantics, the more obvious the difference heat map and line graph between DFGN(reduce) and DFGN(full) is. The dynamic generation mechanism will extract more appropriate semantic information for the matching task.

\subsection{Ablation Analysis}
This section shows the relative validity of the different components of our DFGN(full) model. Table \ref{ablation} presents the results on the TrecQA(clean) test set. We introduce seven different structures. 'w/o' stands for without.

\begin{table}[htbp]
\centering
\caption{Ablation analysis On TrecQA(clean) test set.}
\label{ablation}
  \begin{tabular}{lcc}
    \hline
    Setting & MAP & MRR\\
    \hline
Full Model                    &0.848 &0.928\\
(1)w/o Encoder         &0.790 &0.861 \\
(2)w/o Compare    &0.798 &0.867 \\
(3)w/o Second Co-attention   &0.655 &0.730 \\ 
(4)w/o Intra-attention features  & 0.831 & 0.908\\
(5)w/o Self-attention features   & 0.835 & 0.912\\
(6)w/o Co-attention feature   & 0.833 & 0.908\\
(7)w/o All sentence features & 0.828 & 0.905\\
  \hline
\end{tabular}
\end{table}

(1) We remove the encoder layer, input original embedding and sentence features to co-attention and compare layer. The influence is significantly large, causing MAP to drop by $5.8\%$. This illustrates that the integration of contextual information by encoding is essential to the system.

(2) We get rid of the comparison between co-attention results and encoding output. The MAP drop by $5\%$. It shows that the existence of comparative information is a useful supplement to the model.

(3) We abandon the co-attention encoder before the compare layer, apply encoding results as co-attention compare results. As we expected, lacking second co-attention dramatically reduces the functionality of the entire system. The MAP and MRR reduce more than $19\%$, indicating that interactive information learning is indispensable for sentence matching task.

(4) We take away $4$ sentence features acquired by intra-attention. With MAP decreased by $1.7\%$ and MRR drop by $2.0\%$, intra-attention proves that it extracts the internal information in a sentence.

(5) We withdraw $2$ sentence features extracted by self-attention. The MAP and MRR reduce by $1.3\%$ and $1.6\%$, indicating the parameters in self-attention successfully acquire the extension characteristics of sentences.

(6) We discard $4$ sentence features generated by first co-attention. With MAP decreased by $1.5\%$, interactive features extracted by co-attention prove effective.

(7) We cast aside all $10$ sentence features extracted by multiple attention and pooling strategies, which mean we use only the original embedding. The MAP is cut down by $2\%$. The effectiveness of attaching sentence level information is demonstrated.

From ablation analysis, we can observe the relative functions of various components to our model, and confirm the analysis of section $5.1$ and $5.2$.

\section{CONCLUSION}
We propose a novel architecture Dynamic Feature Generation Network (DFGN) for retrieval-based question answering. Unlike previous work which only focused on feature augmentation in the lexical level, we study dynamic feature extraction and selection in the sentence level. Features are extracted by a variety of different attention mechanisms, attached to the sentence level, and dynamically filtered. DFGN acquire the ability to extract and select features according to different tasks dynamically. Different kinds of characteristics are distilled according to specific tasks, enhancing the practicability and robustness of the model. Our model needs no external resources and feature engineering, relies solely on the semantic information of the text itself. The experimental results outperform current work on multiple well-known datasets, which illustrates our approach effectively improves information retrieval efficiency. Moreover, we give an in-depth analysis of our model which enable us to comprehend its inner working principle further. In the future, we plan to validate the efficiency of our model on more sentence matching tasks, such as natural language inference and paraphrase identification.

